\documentclass[runningheads]{llncs}

\usepackage{eccv}

\usepackage{eccvabbrv}

\usepackage{graphicx}
\usepackage{booktabs}
\usepackage{multirow}
\usepackage{enumitem}
\usepackage[accsupp]{axessibility}  % Improves PDF readability for those with disabilities.

\usepackage[pagebackref,breaklinks,colorlinks,citecolor=eccvblue]{hyperref}
\usepackage{orcidlink}
\usepackage{wrapfig}

\newcommand{\method}{KRONC\xspace}
\newcommand{\dataset}{KRONC-dataset\xspace}
\newcommand{\tit}[1]{\smallskip\noindent\textbf{#1.}}

\usepackage{multirow}
\usepackage[ruled]{algorithm2e}
\usepackage{xcolor}
\usepackage{physics}
\usepackage{colortbl} 
\usepackage{xcolor} 
\usepackage{tikz}

\begin{document}

\title{KRONC: Keypoint-based Robust Camera Optimization
for 3D
% NeRF
%Neural 
Car Reconstruction} 

\titlerunning{KRONC}

\author{Davide Di Nucci \inst{1}\orcidlink{0009-0000-7450-8796} \and
Alessandro Simoni\inst{1}\orcidlink{0000-0003-3095-3294} \and
Matteo Tomei\inst{2}\orcidlink{0000-0002-1385-924X} \\ 
Luca Ciuffreda \inst{2} \and
Roberto Vezzani\inst{1}\orcidlink{0000-0002-1046-6870} \and
Rita Cucchiara \inst{1}\orcidlink{0000-0002-2239-283X}}

\authorrunning{D. Di Nucci et al.}

\institute{University of Modena and
 Reggio Emilia
 \email{\{davide.dinucci,alessandro.simoni,\\roberto.vezzani,rita.cucchiara\}@unimore.it}\\
 \and
Prometeia
 \email{\{matteo.tomei,luca.ciuffreda\}@prometeia.com}}

\maketitle

%\begin{figure}[!t]
%  \centering
%  \includegraphics[height=3.5cm]{figs/vera_fig1.jpeg}

%  \label{fig:example}
%\end{figure}

\begin{abstract}
\vspace{-.7cm}
%The creation of three-dimensional representations of objects or scenes starting from a set of images has been a widely discussed topic for years and has gained additional attention with NeRF-based representations. However, an underestimated prerequisite is the knowledge of camera poses or, more specifically, the estimation of the extrinsic calibration parameters. Although excellent general-purpose Structure from Motion methods are available as a pre-processing step, their computational load is high. This paper introduces \method, a novel approach aimed at inferring view poses by leveraging prior knowledge about the object to reconstruct and its representation through semantic keypoints. 
The three-dimensional representation of objects or scenes starting from a set of images has been a widely discussed topic for years and has gained additional attention after the diffusion of NeRF-based approaches. However, an underestimated prerequisite is the knowledge of camera poses or, more specifically, the estimation of the extrinsic calibration parameters. Although excellent general-purpose Structure-from-Motion methods are available as a pre-processing step, their computational load is high and they require a lot of frames to guarantee sufficient overlapping among the views. This paper introduces \method, a novel approach aimed at inferring view poses by leveraging prior knowledge about the object to reconstruct and its representation through semantic keypoints. 
% In this paper, we propose \method, an innovative technique for estimating the view poses based on some prior knowledge of the object to be reconstructed and on the corresponding ability to represent it using semantic keypoints. 
With a focus on vehicle scenes, \method is able to estimate the position of the views as a solution to a light optimization problem targeting the convergence of keypoints' back-projections to a singular point. To validate the method, a specific dataset of real-world car scenes has been collected. Experiments confirm \method's ability to generate excellent estimates of camera poses starting from very coarse initialization. Results are comparable with Structure-from-Motion methods with huge savings in computation. Code and data will be made publicly available.

\vspace{-0.8cm}
\end{abstract}    
\vspace{-.1cm}

\section{Introduction}
\label{sec:intro}
\vspace{-.1cm}
 \begin{figure}[t]
     \centering
     \includegraphics[page=1,width=0.95\linewidth]{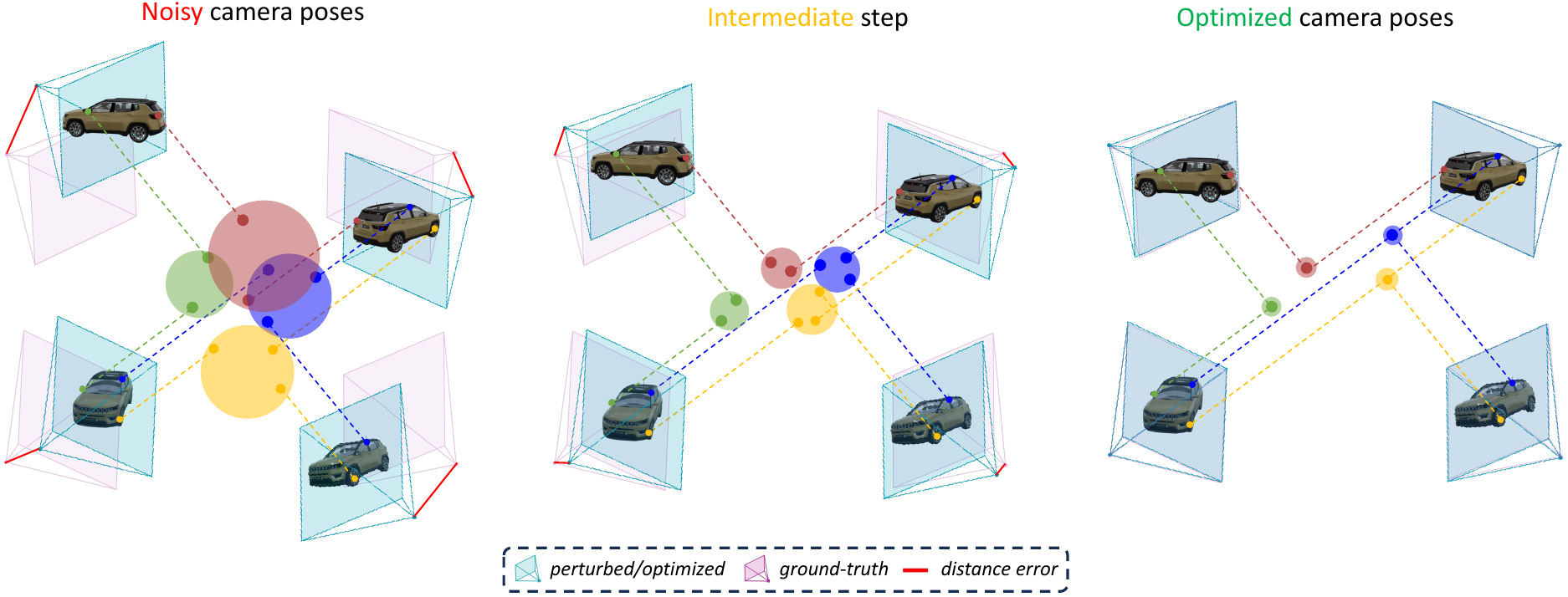}
     \caption{\method is a lightweight camera optimization algorithm for vehicle scenes which leverages 2D semantic keypoints. Keypoints are aligned in a common 3D world reference system, leading to precise camera registration.}
     \label{fig:first_page}
     \vspace{-.5cm}
 \end{figure}

Recent view synthesis techniques, such as Neural Radiance Fields~\cite{mildenhall2021nerf} and 3D Gaussian Splatting~\cite{kerbl20233d}, have revolutionized the reconstruction of both synthetic and real-world scenes. Training only on a few dozen images with known camera poses, they are able to provide high-quality renderings of the scene from novel viewpoints. Their representations emerged as an intermediate domain between the realms of 2D and 3D on which executing standard computer vision tasks such as object detection~\cite{hu2023nerf} or segmentation~\cite{cen2023segment}, paving the way for a variety of applications~\cite{xie2023s, liu2023real, liu2023hosnerf, corona2022mednerf}. For instance, owning a NeRF model and directly applying downstream recognition to it allows for easier inspection and assessment~\cite{buildings13010213}, compared to conducting the same analysis across individual pictures. The \textit{vehicle inspection} task~\cite{di2023carpatch} has recently gained attention for the benefits it can bring to automotive industries and service providers. Its purpose is to generate high-quality renderings of specific car instances from different perspectives starting from a collection of images. This is exactly what NeRF models try to achieve in their broadest formulation, although with a focus on vehicle instances. Facilitating their meticulous inspection without on-site check-up from experts could be extremely convenient for car manufacturers to determine eventual external defects, for insurance companies to estimate post-accident damages and repair costs, or for car rentals for liability assessment automation.

% This paper specifically tackles the \textit{vehicle inspection} task through novel view synthesis algorithms, by observing that: 
However, applying standard novel view synthesis approaches to vehicle reconstruction highlights the following limitations:
(i) recent NeRF and Gaussian Splatting methods still rely on classical Structure-from-Motion pipelines (\eg~COLMAP~\cite{schonberger2016structure}) for camera parameters estimation, sometimes even exceeding the time and resource requirements of the actual downstream optimization~\cite{muller2022instant, sun2022direct};
% (i) recent NeRF methods achieve nearly instantaneous optimization of radiance fields, demonstrating state-of-the-art performance; 
(ii) to the best of our knowledge, no dedicated datasets are available for comprehensive real-world vehicle reconstruction, with the evaluation still limited to synthetic scenes.
Moreover, vehicles represent a well-studied and deeply modeled object category in the computer vision literature (for \eg~large-scale unbounded scene recognition and autonomous driving~\cite{geiger2012we, song2019apollocar3d, dosovitskiy2017carla}), leading to a pool of established works and priors to be leveraged for \textit{vehicle inspection}, too.

% Among the others, generating high-quality car renderings from different perspectives has recently gained attention~\cite{di2023carpatch}, striving to facilitate a meticulous inspection of vehicles post-accidents, thereby streamlining insurance and rental applications. This paper specifically tackles the \textit{vehicle inspection} task through novel view synthesis algorithms, by observing that: (i) recent NeRF methods achieve nearly instantaneous optimization of radiance fields, demonstrating state-of-the-art performance~\cite{muller2022instant, sun2022direct}; (ii) most of them still rely on classical Structure from Motion (S\textit{f}M) or Simultaneous Localization and Mapping (SLAM) pipelines for camera parameters estimation, often exceeding the time and resource requirements of the neural radiance field optimization; (iii) to the best of our knowledge, no dedicated datasets are available for high-fidelity and comprehensive real-world vehicle reconstruction.
Motivated by these observations, in this paper we propose an efficient algorithm for camera frame registration, which is able to break the dependency on heavy COLMAP-like pre-processing. Moreover, we release a new benchmark (the \dataset) of real-world vehicle scenes, with the aim of fostering novel view synthesis for \textit{vehicle inspection}.
To avoid S\textit{f}M, recent works proposed Bundle-Adjusting NeRF~\cite{lin2021barf, chen2023local} by jointly reconstructing neural fields and registering camera frames. Their benefits come at the cost of integrating camera alignment in neural field optimization, which is not as straightforward as performing the two steps sequentially. We show that comparable performance can be obtained for vehicle reconstruction by keeping the two steps separated, exploiting a much lighter alternative to raw RGB pixels for camera optimization, \ie 2D keypoints, making computational overhead negligible. Our proposal combines the efficiency of bundle adjustment and the flexibility of stand-alone S\textit{f}M packages, making it suitable for every downstream novel view synthesis technique not limited to NeRFs. As shown in Fig.~\ref{fig:first_page}, our \method algorithm projects semantically consistent keypoints from multiple views to a common 3D world's reference system and pushes them close together. Doing so, it tries to figure out both a reasonable configuration of cameras and meaningful depths for keypoints. Differently from incremental S\textit{f}M and methods relying on pairwise image correspondences~\cite{truong2023sparf, lao2024corresnerf}, \method conducts global alignment, optimizing absolute camera positions depending on semantic keypoints shared between all viewpoints (without needing any matching algorithm).

On synthetic vehicle scenes our results show improved performance w.r.t. state-of-the-art bundle-adjustment methods, by adding the same camera noise to ground truth poses and attempting to restore a coherent disposition. On real scenes from the \dataset, we captured cars with mobile devices by performing a full 360° counterclockwise rotation around the car, which is the standard way of capturing scenes for large object reconstruction~\cite{nerfstudio}. State-of-the-art bundle adjustment solutions struggle to converge in this setting. By coarsely initializing the poses of the cameras following a simple handcrafted circular trajectory, our keypoint-based registration method is able to find a good camera arrangement even when reducing the number of input images by 75\%, while COLMAP performance rapidly drops. 
To sum up, our contributions encompass the following:
\begin{itemize}[topsep=0pt]
    \item We present the \dataset of real-world, high-quality car scenes, specifically devised for novel view synthesis in the context of \textit{vehicle inspection}.
    \item We introduce an efficient keypoint-based camera registration (\method) algorithm to be executed before neural radiance field optimization, keeping the two separate steps and allowing for higher flexibility compared to bundle-adjusting NeRFs from noisy cameras.
    \item On real scenes, we leverage the typical behavior of capturing a scene by making cameras follow a circular trajectory, recovering a plausible pose configuration with a speedup reaching one order of magnitude \wrt COLMAP.
\end{itemize}
\section{Related work}
\label{sec:related}

In this section, we provide an overview of methodologies centered on camera pose estimation and 3D reconstruction techniques.

\tit{Novel view synthesis} The success of NeRF~\cite{mildenhall2021nerf} resulted in follow-up strategies aiming to improve both quality and speed. Mip-NeRF~\cite{barron2021mip, barron2022mip} along with Zip-NeRF~\cite{barron2023zip} reach state-of-the-art novel-view generation quality by introducing anti-aliasing procedures. A widely recognized issue in 3D reconstruction concerns computation. Several efforts have demonstrated the viability of achieving high-fidelity reconstructions while also shortening overall training time. Instant-NGP~\cite{muller2022instant} employs a multi-resolution hash table alongside a streamlined MLP architecture, enhancing the efficiency of the training process. Alternative techniques like DVGO~\cite{sun2022direct} focus on optimizing voxel grids containing features to facilitate rapid reconstruction of radiance fields. TensoRF~\cite{chen2022tensorf} combines the traditional CP decomposition with a novel vector-matrix decomposition technique~\cite{carroll1970analysis} resulting in accelerated training and improved reconstruction quality. Apart from neural radiance fields optimization, other approaches like Gaussian Splatting~\cite{kerbl20233d} have demonstrated impressive outcomes by characterizing the 3D space as a collection of Gaussians. To meet the real-time requirements of \textit{vehicle inspection}, we choose baselines based on training time $\vs$ reconstruction quality trade-offs.

\tit{Bundle-adjustment and pose refinement}
Estimating or refining camera poses represents a critical challenge in both NeRFs and vehicle inspection domains. Structure-from-Motion (S\textit{f}M) techniques~\cite{agarwal2011building, hartley2003multiple, pollefeys1999self, pollefeys2004visual, schoenberger2016mvs, snavely2006photo, snavely2008modeling} are widely established methods for acquiring precise geometry and camera poses from video or image data through an offline per-scene optimization process.
In contrast, simultaneous localization and mapping (SLAM) methods~\cite{campos2021orb, engel2014lsd, mur2015orb}, typically operate online. However, they are known to exhibit unreliability in scenarios with heavily rotating trajectories or scenes containing sparse visual features.
Works such as GNeRF~\cite{meng2021gnerf}, NeRF++~\cite{wang2021nerf}, and SinERF~\cite{xia2022sinerf} made efforts towards enhancing the camera poses within NeRF architectures.
Approaches such as Barf~\cite{lin2021barf} and L2G-NeRF~\cite{chen2023local} employ a joint optimization strategy to refine both the radiance field and camera parameters starting from noisy poses. These methods rely solely on the photometric loss as the training signal during optimization.
Newer techniques, such as Sparf~\cite{truong2023sparf}, KeypointNeRF~\cite{mihajlovic2022keypointnerf} and CorresNeRF~\cite{lao2024corresnerf} aim to enhance camera pose estimation in few-images scenarios by leveraging multi-view correspondences derived from matches between training views. 
However, they depend on pairwise image correspondences. In contrast, we introduce a novel method for refining poses based on keypoint information shared among multiple views.

\tit{Datasets}
The NeRF Synthetic Blender dataset~\cite{mildenhall2021nerf} is one of the most extensively used benchmarks for assessing NeRFs performance. This dataset consists of scenes created with Blender\footnote{\url{http://www.blender.org}}.
Other synthetic datasets include the Blend DMVS~\cite{yao2020blendedmvs}, which provides scenes at different scales, and the Shiny Blender dataset~\cite{verbin2022ref}, which mostly contains objects with simple geometries.
Regarding vehicle inspection, the only real-world resource holding captures from a single high-quality vehicle has been introduced in Ref-NeRF~\cite{verbin2022ref}. While datasets like Tanks and Temples~\cite{Knapitsch2017} and LLFF~\cite{mildenhall2019llff} serve as valuable benchmarks for evaluating novel view synthesis in various real-world scenarios, their scope might not be comprehensive enough for in-depth studies focused on vehicles.
The CarPatch~\cite{di2023carpatch} dataset, despite its detailed annotations and scene diversity, provides synthetic cars only. Our proposed \dataset aims to address these limitations by facilitating the evaluation of cars from real-world settings, by filling a crucial gap in evaluating and improving vehicle inspection.

\definecolor{myrowcolor}{HTML}{F5F5F5}
\definecolor{tabfirst}{HTML}{dec546}
\definecolor{tabsecond}{HTML}{d7d7d7}
\definecolor{tabthird}{HTML}{7e4205}

\section{The \dataset}
\label{sec:dataset}

In this section, we discuss the source data and the methodology employed to create our \dataset. Specifically, motivated by the lack of data pertinent to vehicle inspection within real-world settings, we detail the steps carried out to manually gather car scenes.

\begin{table}[t]
    \centering
    \caption{Summary of the \dataset: for each scene, we report the vehicle model, the number of images, and the average number of keypoints per image.}
    \resizebox{0.9\linewidth}{!}{
    \begin{tabular}{lcccc ccc c}
        \toprule
         \textbf{Env} & \textbf{Env1} & \textbf{Env1} & \textbf{Env1} & \textbf{Env2} & \textbf{Env2} & \textbf{Env3} & \textbf{Env3} \\
         \midrule
         \textbf{Vehicle} & Ford-Focus & Fiat-500L & Hyundai-i10 & Fiat-500L & Toyota-Yaris & Toyota-Yaris & Hyundai-i10 \\
         \rowcolor{myrowcolor}
         \textbf{Images} & 161 & 143 & 123 & 94 & 91 & 116 & 123 \\
         \textbf{\#Kpts} & 23 & 14 & 19 & 13 & 20 & 22 & 16 \\
         \bottomrule
    \end{tabular}}
    \label{tab:tab_kronc_dataset}
    \vspace{-0.5cm}
\end{table}

% \subsection{Enhancing Vehicle Inspection Insights} 
% Benchmark datasets like Tanks and Temples~\cite{Knapitsch2017}, LLFF~\cite{mildenhall2019llff}, and the Unbounded 360° dataset~\cite{barron2022mip} have played pivotal roles, serving as standards to evaluate the efficacy of cutting-edge NeRF (Neural Radiance Fields) methods in real-world scenarios. These datasets, widely referenced in literature, have significantly contributed to advancing the understanding and development of computational methods for scene reconstruction and rendering.
% However, these datasets, while efficient, lack specific information pertinent to vehicle inspection within real-world settings. Recognizing this significant gap, we introduce the KRONC dataset as an innovative solution tailored explicitly to address these challenges.  \\

\subsection{Dataset captures} 
The dataset has been collected by employing different devices in three distinct environments. For the first two environments (Env1 and Env2), the scenes have been captured using two standard smartphone cameras (OnePlus 7T and OnePlus Nord). For Env3, we adopted a DJI MINI 2 SE drone for taking pictures. Three different scenes belong to Env1 and two additional scenes come from Env2 and Env3, respectively, leading to a total of 7 scenes. Each scene represents a single vehicle captured from multiple viewpoints. To mimic user behavior in real use cases, we opted for capturing video clips by moving around the vehicle, following a circular path around each car, while maintaining a consistent distance throughout the registration. In each video, a single complete lap around the car has been performed, making the last frame roughly correspond to the first one. Each capture was intended to include the entire car body in the field of view of the camera. Note that this represents the suggested way of capturing large bounded objects even from well-known 3D reconstruction services\footnote{\url{https://lumalabs.ai/}}.

Original videos have been captured with a frame rate ranging from 30 to 60 fps, before being downsampled to 5 fps.
%using FFmpeg. 
Frames have been extracted and sub-sampled again to make data suitable for S\textit{f}M pipelines and novel view synthesis processing. Both the original videos and the selected frames are available to download inside the public dataset for completeness and future fair comparisons. 

\subsection{Dataset metadata}
To leverage car keypoints for camera extrinsic optimization (as will be detailed in Sec.~\ref{sec:method}), we automatically annotated semantic keypoints on each single frame of the \dataset, by adopting the OpenPifPaf~\cite{kreiss2021openpifpaf} framework. Specifically, we used the ShufflenetV2K16 model~\cite{ma2018shufflenet} trained to predict the 66 distinct keypoints defined in ApolloCar3D~\cite{song2019apollocar3d}. Moreover, for vehicle inspection purposes, we provide car instance segmentation masks to make it possible to discard unnecessary background pixels. Image-wise mask predictions have been obtained through Mask2Former~\cite{cheng2021mask2former} with a Swin Large~\cite{liu2021swin} backbone trained on the
COCO panoptic dataset~\cite{kirillov2019panoptic}. Masks isolate the vehicle from complex backgrounds, allowing to focus on vehicle reconstruction in presence of challenging environmental conditions, which however is not the case for the \dataset.
Finally, each dataset underwent rigorous COLMAP~\cite{schoenberger2016mvs} processing to estimate precise camera poses. This information can be used as an upper-bound reference for evaluating pose estimation methods, highlighting the remarkable precision achieved by COLMAP, especially when large volumes of images are available.
Table \ref{tab:tab_kronc_dataset} presents a summary of the scenes included in the \dataset along with the corresponding number of images per scene and the average number of keypoints detected per image.

\section{Keypoint-based camera optimization}
\label{sec:method}

\setlength{\belowdisplayskip}{2pt} \setlength{\belowdisplayshortskip}{2pt}
\setlength{\abovedisplayskip}{2pt} \setlength{\abovedisplayshortskip}{2pt}
In this section, we detail how vehicle semantic keypoints can benefit multi-view consistency and camera pose alignment, as a pre-processing step to improve downstream novel views synthesis algorithms (\eg Neural Radiance Fields~\cite{mildenhall2021nerf}).

\subsection{Exploiting keypoint projections}
\label{subsec:keypoints_2d_to_3d}
The input of our algorithm is a set of $N$ captures $\mathcal{I}=\{I_i\}_{i=1}^N$ of a scene representing a vehicle. 
Without loss of generality, we assume that the $N$ images have been taken with the same camera, whose internal calibration parameters are known or have been previously calculated. Therefore, we can define a unique matrix $K \in \mathbb{R}^{3 \times 3}$ containing the intrinsic parameters, common to all the views.

Let $R_i \in SO(3)$, $\mathbf{t}_i \in \mathbb{R}^{3}$, be the extrinsic parameters (\ie, rotation matrix and translation vector) of each image $I_i$ with respect to a common world reference system. For images captured with a moving camera, these parameters are generally not available and should be estimated with computationally-intensive procedures such as S\textit{f}M algorithms (\eg COLMAP~\cite{schonberger2016structure}). \method optimizes a noisy/coarse initial approximation of the extrinsic camera parameters. Differently from recent methods exploiting visual pairwise image correspondences~\cite{truong2023sparf, lao2024corresnerf}, we benefit from a much lighter global information shared between (potentially) all the captures, \ie semantic 2D keypoint coordinates. 

\tit{Projecting keypoints to the 3D world}
Let $\{p^1, p^2,...,p^J\}$ be a set of $J$ semantic keypoints, meaningful for a class of interesting objects (vehicles, in our scenario). Each input image is required to be annotated with the 2D position of these keypoints. The estimation of the 2D keypoint coordinates is a common task in computer vision~\cite{song2019apollocar3d} and the corresponding algorithm remains outside the scope of this work. 
Therefore, let us define the available set of keypoints as 
%) 
$\mathcal{P} = \{p^j_i\}$, 
$p^j_i=(u_i^j,v_i^j,m_i^j,z_i^j)$, 
where $(u_i^j,v_i^j)$ are the 2D coordinates of the $j$-th keypoint in the $i$-th image plane, $m_i^j \in [0,1]$ is the visibility of the keypoint and $z_i^j$ is the distance of the keypoint from the camera center. We introduce $m_i^j$ as a consequence of potential occlusions, since we may observe only a subset of the $J$ keypoints in each image. However, we assume that the number of views $N$ is large enough to guarantee a certain degree of overlap between views, resulting in the same semantic keypoint $p^j$ being visible in multiple captures. The additional $z_i^j$ is required to back-project $p_i^j$ from the 2D image plane to a common 3D world's reference frame $XYZ$ as follows:

\begin{equation}
    \begin{bmatrix}
        X_i^j\\
        Y_i^j\\
        Z_i^j\\
    \end{bmatrix}
    =
    \setlength\arraycolsep{2pt}
    \begin{bmatrix}
    R_i & \mathbf{t}_i
    \end{bmatrix}
    \setlength\arraycolsep{0.5pt}
    \begin{bmatrix}
    & K^{-1} &\\
    0 & 0 & 1
    \end{bmatrix}
    \begin{bmatrix}
        u_i^j\\
        v_i^j\\
        1\\
    \end{bmatrix}
    z_i^j.
\label{eq:kp_2d_to_3d}
\end{equation}

Since both camera parameters $R_i$, $\mathbf{t}_i$ and keypoint's depth $z_i^j$ in the camera's reference system are unknown or initialized with some noisy values, we need to find a suitable procedure to optimize them. In Sec.~\ref{sec:experiments} we detail how these parameters are initialized for both synthetic and real vehicle scenes. In the remaining of this section, we describe how we optimize camera poses and keypoints' depths to ensure 2D re-projection consistency between captures.

\subsection{3D centroids and re-projection consistency}
\label{subsec:keypoints_3d_to_2d}
The optimization of the camera poses is based on the following assumption: the 3D back-projections of the same semantic keypoint $p^j$ from different views should lie on the same 3D point. However, if the extrinsic parameters and depths are affected by noise, a cluster of 3D points will be generated for a specific semantic keypoint. We aim to align each back-projected semantic keypoint $p^j$ with its cluster's centroid. Taking into account a specific view, its extrinsic parameters will be optimal when the distances of its back-projected keypoints from the corresponding cluster centers are minimized. The same holds for each keypoint depth $z_i^j$.
In our preliminary experiments, we empirically observed better results and convergence by minimizing the Euclidean distance between each keypoint and its cluster center both after re-projecting them onto each image plane and directly in the 3D space.

%For a given semantic keypoint $p^j$, our camera optimization algorithm endeavors to minimize the distance between $p_i^j$ in captures $I_i$ and the centroid of the cluster formed by the corresponding keypoints $p^j$ across all the captures in which it is visible, after projecting them to the 3D world. In our preliminary experiments, we observed better results and smoother convergence when re-projecting the cluster centroid onto the views $I_i$, and calculating the distances on the 2D image planes, as opposed to directly operating in 3D. Details of the KRONC mechanism for camera optimization is presented in Algorithm~\ref{alg:kronc}.

\tit{3D clusters and centroids re-projection}
Formally, let's consider a semantic keypoint $p^j$ at a time. Let $M^j$ be the number of images where the $j$-th keypoint is visible, \ie $M^j = \sum_i{m_i^j}$.
We independently project all the keypoints $p_i^j$ from these images to the common 3D world reference frame through Eq.~\ref{eq:kp_2d_to_3d}, before computing their 3D centroid $C^j$ as follows:

% \begin{equation}
%     C^j = (\overline{x}^{C^j}, \overline{y}^{C^j}, \overline{z}^{C^j}) = \\
%     \left(\sum_{i=1}^{J}\frac{\overline{x}_i^j}{J}, \sum_{i=1}^{J}\frac{\overline{y}_i^j}{J}, \sum_{i=1}^{J}\frac{\overline{z}_i^j}{J}\right).
% \label{eq:kp_centroid_3d}
% \end{equation}

\begin{equation}
    C^j = 
    \begin{bmatrix}
        X_C^j\\
        Y_C^j\\
        Z_C^j
    \end{bmatrix}
    = \frac{1}{M^j} \sum_{i} { \left(
    m_i^j \cdot
    \begin{bmatrix}
        X_i^j\\
        Y_i^j\\
        Z_i^j
    \end{bmatrix} \right)}.
\label{eq:kp_centroid_3d}
\end{equation}

The 3D cluster's centroid $C^j$ can be re-projected into each $i$-th image $I_i$ and compared to the corresponding annotated keypoint (if visible). The coordinates $(u^j_{C,i}, v^j_{C,i})$ of the re-projected centroid can be computed as follows:

\begin{equation}
    \begin{bmatrix}
        u^j_{C,i}\\
        v^j_{C,i}\\
        1
    \end{bmatrix}
    \propto
    \setlength\arraycolsep{2pt}
        \begin{bmatrix}
        & 0 \\
        K & 0 \\
        & 0
        \end{bmatrix}
    \setlength\arraycolsep{2pt}
    \begin{bmatrix}
    & R_i & \mathbf{t}_i & \\
    0 & 0 & 0 & 1
    \end{bmatrix}^{-1}
    \begin{bmatrix}
        X_C^j\\
        Y_C^j\\
        Z_C^j\\
        1
    \end{bmatrix}.
\label{eq:centroid_3d_to_2d}
\end{equation}

%RV: si usa spesso il propto invece di = 
%and we convert back from homogeneous to Euclidean coordinates by $z_i^{C^j}$ normalization.

%Our optimization objective for the $j^{th}$ keypoint within the $i^{th}$ capture aims to minimize the Euclidean distance between the original keypoint's coordinates $(x_i^j, y_i^j)$ and the coordinates of the re-projected cluster's centroid $(x_i^{C^j}, y_i^{C^j})$:

For each image and for each visible keypoint, we aim to minimize the following optimization objective:
%\begin{equation}
%    \mathcal{L}_i^j(R_i, \mathbf{t}_i, z_i^j) = \sqrt{(x_i^j-x_i^{C^j})^2+(y_i^j-y_i^{C^j})^2}.
%\label{eq:img_objective}
%\end{equation}
\begin{equation}
    % \mathcal{L}_i^j(R_i, \mathbf{t}_i, z_i^j) = \| (u_i^j, v_i^j), (u^j_{C,i}, v^j_{C,i}) \|_2.
    \mathcal{L}_i^j(R_i, \mathbf{t}_i, z_i^j) = \| (X_i^j, Y_i^j, Z_i^j)-(X_C^j, Y_C^j, Z_C^j) \|_2 + \lambda\| (u_i^j, v_i^j)-(u^j_{C,i}, v^j_{C,i}) \|_2
\label{eq:img_objective}
\end{equation}
where $\lambda$ balances the magnitude of distances in the 3D world (as meters) and distances on the image plane (as pixels).

\tit{Full optimization objective}
The algorithm seeks to find the global minimum of the following loss, by concurrently optimizing keypoint's projections and back-projection for all the captures $\mathcal{I}$ and for all the semantic keypoints $\mathcal{P}$:
\begin{equation}
    \min_{R_i,\mathbf{t}_i,z_i^j}\frac{1}{J}\sum_{j=1}^{J}\sum_{i=1}^{N}m^j_i\mathcal{L}_i^j(R_i, \mathbf{t}_i, z_i^j).
\label{eq:full_objective}
\end{equation}

\begin{algorithm}[t]
% \small
    \SetKwInOut{Input}{Input}
    \SetKwInOut{Output}{Output}
    \SetKwInOut{Params}{Params}
    \Input{Images $\mathcal{I}=\{I_i\}_{i=1}^N$, \\
    semantic keypoints $\mathcal{P} = \{p^j\}_{j=1}^J$ \\
    visibility $m^j_i$ of keypoint $p^j$ on image $I_i$ \\
    noisy $R_i$, $\mathbf{t}_i$, $z_i^j$, defining $\pi_i$ projection, \\ 
    function $f$ mapping $R_i \in \mathbb{R}^{3 \times 3}$ to $\mathbf{r}_i \in \mathbb{R}^{6}$; \\
    }
    %2D-to-3D mapping $\pi_i$ (Eq.~\ref{eq:kp_2d_to_3d}), \\
    %$i \in \{ 1, ..., N \}$, $j \in \{ 1, ..., J \}$
    %;}
    \Output{Optimized $R_i$, $\mathbf{t}_i$, $z_i^j$;}
    \Params{number of steps $S$, \\
    learning rate $\eta$, \\
    2D loss weight $\lambda$; }
    $\mathbf{r}_i=f(R_i)$ \;
    \For{$s:=1 \to S$}{
        $R_i = f^{-1}(\mathbf{r}_i)$ \;
        $\mathcal{L}=0$ \;
        \ForEach{$p^j \in \mathcal{P},~j \in \{ 1, ..., J \}$}{
            $C^j = \frac{1}{\sum_{i=1}^{N}m_i^j}\sum_{i=1}^{N}m_i^j\pi_i(p_i^j)$ \;
            \ForEach{$I_i \in \mathcal{I},~i \in \{ 1, ..., N \}$}{
                $\mathcal{L} = \mathcal{L} + m^j_i\left(\lVert \pi_i(p_i^j)-C^j \rVert_2 + \lambda\lVert p_i^j-\pi_i^{-1}(C^j)\rVert_2\right)$ \;
            }
        }
    $\mathbf{r}_i = \mathbf{r}_i - \eta\pdv{\mathcal{L}}{\mathbf{r}_i},~i \in \{ 1, ..., N \}$ \;
    $\mathbf{t}_i = \mathbf{t}_i - \eta\pdv{\mathcal{L}}{\mathbf{t}_i},~i \in \{ 1, ..., N \}$ \;
    $z_i^j = z_i^j - \eta\pdv{\mathcal{L}}{z_i^j},~i \in \{ 1, ..., N \},~j \in \{ 1, ..., J \}$ \;
    }
\caption{KRONC algorithm. Note that \textbf{foreach} statements here represent parallel operations in our implementation }
\label{alg:kronc}
\end{algorithm}

Although no constraints limit the direct optimization of translation embeddings $\mathbf{t}_i \in \mathbb{R}^{3}$ and depth values $z_i^j \in \mathbb{R}$, the same does not hold for rotation matrices, which must preserve orthogonality. Inspired by recent works facing the same issue~\cite{truong2023sparf, jeong2021self}, we adopt the 6D representation of~\cite{zhou2019continuity}, where the unnormalized first two columns of the rotation matrix are employed to represent a full rotation. Specifically, given the noisy rotation matrix for the $i$-th image $R_i = [\mathbf{a}_1 \,\, \mathbf{a}_2 \,\,  \mathbf{a}_3] \in \mathbb{R}^{3 \times 3}$, we compute the corresponding initial rotation vector $\mathbf{r}_i=[\mathbf{a}_1^T,\mathbf{a}_2^T] \in \mathbb{R}^{6}$ by simply dropping the last column. At every optimization step, we first recover the full rotation matrix as $R_i = [\mathbf{b}_1 \,\, \mathbf{b}_2 \,\,  \mathbf{b}_3] \in \mathbb{R}^{3 \times 3}$, where $\mathbf{b}_1=N(\mathbf{a}_1)$, $\mathbf{b}_2=N(\mathbf{a}_2-(\mathbf{b}_1 \cdot \mathbf{a}_2)\mathbf{b}_1)$, $\mathbf{b}_3=\mathbf{b}_1\times\mathbf{b}_2$, and $N$ denotes L2 normalization. Then, we compute our objective and update $\mathbf{r}_i$, $\mathbf{t}_i$ and $z_i^j$ according to Eq.~\ref{eq:full_objective}.

The optimization is carried out through several iterations. At each iteration, the new positions of the cluster centers are concurrently computed and the parameters are optimized in parallel using gradient descent, leading to almost real-time optimization on the latest GPU devices.

The KRONC algorithm is devised as an easy-to-implement and efficient camera alignment strategy to be executed before novel view synthesis methods. Note that it does not make any use of the raw RGB image values, but only exploits keypoints projections from 2D to 3D and vice versa. It does not jointly optimize for neural 3D representations and camera registration as other methods do~\cite{lin2021barf, chen2023local}, allowing for seamless integration with every downstream method requiring accurate camera poses. KRONC is detailed in Algorithm~\ref{alg:kronc}.

% \tit{Camera extrinsic parameterization}

\section{Experimental evaluation}
\label{sec:experiments}

In this section, we present the experimental settings and the results obtained using \method for camera registration, followed by different state-of-the-art downstream novel view synthesis approaches. Performances are evaluated on synthetic and real-world vehicle scenes. 
In accordance with Barf~\cite{lin2021barf}, we apply Procrustes analysis to determine a 3D similarity transformation for aligning the optimized poses with the ground truth, before computing rotation and translation errors $\epsilon_{R}$ and $\epsilon_{\mathbf{t}}$, respectively.
For novel view synthesis evaluation, we adopt common visual quality metrics, \ie~PSNR, SSIM~\cite{wang2004image}, and LPIPS~\cite{zhang2018unreasonable}.

\subsection{Synthetic vehicle scenes}
\label{subsec:experiments_synthetic}
We use the CarPatch dataset~\cite{di2023carpatch} as our benchmark for synthetic 3D vehicle reconstruction evaluation. We adopt the full version containing 8 scenes, each comprising 100 training and 200 test images with ground truth camera poses. Since \method requires the annotation of keypoints, we added them in the original CarPatch 3D Blender models, following the semantic convention defined in~\cite{song2019apollocar3d}. Then, we enriched the CarPatch scenes with ground truth 2D vehicle keypoints via Blender rendering. CarPatch keypoint annotations will be released together with the \dataset.

\definecolor{myrowcolor}{HTML}{F5F5F5}
\definecolor{first}{HTML}{dec546}
\definecolor{second}{HTML}{d7d7d7}
\definecolor{third}{HTML}{7e4205}

\begin{table*}[t]
    \centering
        \caption{Quantitative results averaged over the CarPatch scenes. We assign gold, silver, and bronze medals to the best three methods. }
    \resizebox{\linewidth}{!}{
    \begin{tabular}{lc c|c cc c|c ccc r}
        \toprule
         \textbf{Method} & \textbf{Poses} & & &  $\epsilon_{R} (^{\circ})~\downarrow$ & $\epsilon_{\mathbf{t}}$ (cm)~$\downarrow$ & & &  PSNR~$\uparrow$ & SSIM~$\uparrow$ & LPIPS~$\downarrow$ & Runtime \\
         \midrule
         TensoRF~\cite{chen2022tensorf} & GT & & & - & - & & & 
         34.74 \tikz\draw[second,fill=second,opacity=0.,fill opacity=0.](0,0)circle(.5ex); & 
         0.973 \tikz\draw[second,fill=second,opacity=0.,fill opacity=0.](0,0)circle(.5ex); & 
         0.043 \tikz\draw[second,fill=second,opacity=0.,fill opacity=0.](0,0)circle(.5ex); & 
         35 min \\
         \rowcolor{myrowcolor}
         DVGO~\cite{sun2022direct} & GT & & & - & - & & & 
         36.09 \tikz\draw[second,fill=second,opacity=0.,fill opacity=0.](0,0)circle(.5ex); & 
         0.979 \tikz\draw[second,fill=second,opacity=0.,fill opacity=0.](0,0)circle(.5ex); & 
         0.024 \tikz\draw[second,fill=second,opacity=0.,fill opacity=0.](0,0)circle(.5ex); & 
         10 min \\
         GaussianSplatting~\cite{kerbl20233d} & GT & &  & - & - & & & 
         34.86 \tikz\draw[second,fill=second,opacity=0.,fill opacity=0.](0,0)circle(.5ex); & 
         0.982 \tikz\draw[second,fill=second,opacity=0.,fill opacity=0.](0,0)circle(.5ex); & 
         0.014 \tikz\draw[second,fill=second,opacity=0.,fill opacity=0.](0,0)circle(.5ex); & 
         5 min \\
         \midrule
         \rowcolor{myrowcolor}
         Barf~\cite{lin2021barf} & Noisy & &  &
          7.67 \tikz\draw[third,fill=third](0,0)circle(.5ex); & \hspace{-0.84em}
         49.38  \tikz\draw[third,fill=third](0,0)circle(.5ex); & & & 
         17.46 \tikz\draw[second,fill=second,opacity=0.,fill opacity=0.](0,0)circle(.5ex); & 
         0.870 \tikz\draw[second,fill=second,opacity=0.,fill opacity=0.](0,0)circle(.5ex); & 
         0.142 \tikz\draw[second,fill=second,opacity=0.,fill opacity=0.](0,0)circle(.5ex); & 
         12 h \\ 

                  L2G-NeRF~\cite{chen2023local} & Noisy & &  & 
         0.50 \tikz\draw[first,fill=first](0,0)circle(.5ex);&
         5.26 \tikz\draw[second,fill=second](0,0)circle(.5ex); & & & 
         31.91 \tikz\draw[first,fill=first,opacity=0.,fill opacity=0.] (0,0) circle (.5ex); & 
         0.966 \tikz\draw[second,fill=second,opacity=0.,fill opacity=0.](0,0)circle(.5ex); & 
         0.060 \tikz\draw[second,fill=second,opacity=0.,fill opacity=0.](0,0)circle(.5ex); & 
         6 h \\
         \midrule
         \rowcolor{myrowcolor}
         KRONC + TensoRF~\cite{chen2022tensorf} & Noisy & & & 
         0.65 \tikz\draw[second,fill=second](0,0)circle(.5ex); & 
         3.06 \tikz\draw[first,fill=first](0,0)circle(.5ex); & & & 
         33.80 \tikz\draw[third,fill=third] (0,0) circle (.5ex); & 
         0.971 \tikz\draw[second,fill=third](0,0)circle(.5ex); & 
         0.042 \tikz\draw[third,fill=third](0,0)circle(.5ex); & 
         35.5 min \\
         KRONC + DVGO~\cite{sun2022direct} & Noisy & & & 
         0.65 \tikz\draw[second,fill=second](0,0)circle(.5ex); &
         3.06  \tikz\draw[first,fill=first](0,0)circle(.5ex);& & & 
         34.03 \tikz\draw[second,fill=second](0,0)circle(.5ex); & 
         0.975 \tikz\draw[second,fill=second](0,0)circle(.5ex); & 
         0.029 \tikz\draw[second,fill=second](0,0)circle(.5ex); & 
         10.5 min \\
         \rowcolor{myrowcolor}
         KRONC + GaussianSplatting~\cite{kerbl20233d} & Noisy & & & 
         0.65  \tikz\draw[second,fill=second](0,0)circle(.5ex);& 
         3.06  \tikz\draw[first,fill=first](0,0)circle(.5ex); & & & 
         34.38 \tikz\draw[first,fill=first](0,0)circle(.5ex); & 
         0.982 \tikz\draw[first,fill=first](0,0)circle(.5ex); & 
         0.014 \tikz\draw[first,fill=first](0,0)circle(.5ex); & 
         5.5 min \\
         \bottomrule
    \end{tabular}}

    \label{tab:tab_synt_results}
    \vspace{-.5cm}
\end{table*}

\tit{Implementation and experimental settings} We parametrize the camera poses with the SE(3) Lie algebra and assume known intrinsics. According to the Lego dataset setting of L2G~\cite{chen2023local}, we synthetically perturb the camera poses creating noisy $R\mathbf{t}$ matrices. Noise values for $R$ and $\mathbf{t}$ are sampled from normal distributions with standard deviation $\sigma_{R}=4^{\circ}$ and $\sigma_{\mathbf{t}}=0.5$ m, respectively.
Similarly to COLMAP, we optimize the test poses together with train poses during camera optimization. This differs from L2G and Barf settings, where they perform test-time photometric pose optimization~\cite{lin2019photometric, yen2021inerf} before evaluating view synthesis quality. Given the different ground truth camera distribution between the test set and the training set, we chose to partition each scene of the CarPatch training set into $80$ images for training and $20$ for testing. For an early plausible 3D keypoint back-projection (Eq.~\ref{eq:kp_2d_to_3d}), we randomly initialize the $z^j_i$ values from the range $[\frac{1}{2}\omega, \omega]$, where $\omega$ is the average L2 norm of the translation vectors of the initial camera poses in the scene. Different initialization methods are explored later in Sec.~\ref{subsec:experiments_real}.
As our method is designed to be plug-and-play, we demonstrate its versatility by evaluating the effect of optimized poses on various downstream novel view synthesis methods without modifying their original implementations. 
When selecting novel view synthesis architectures, we were driven by the best trade-off between training time and reconstruction quality, with the goal of developing a real-time system tailored for vehicle inspection.
All the experiments are conducted using a single GeForce GTX 1080 Ti. For consistency, input resolution is fixed to $400\times400$, as in L2G and Barf experimental settings.
After a comprehensive assessment of various methods, we select the following baselines:
\begin{itemize}
 \item \textbf{Gaussian Splatting~\cite{muller2022instant}}: the experiments are conducted without altering the original settings. We train for $10$k iterations before rendering test images.
 \item \textbf{TensoRF~\cite{chen2022tensorf}}: we choose to employ the Nerfstudio~\cite{nerfstudio} implementation for TensoRF. Our configuration involves a batch size of $4096$ rays, a scale dimension of $0.5$, and an initial learning rate set to $0.0001$ with an exponential decay scheduler. Training lasts $10$k iterations.
\item \textbf{DVGO~\cite{sun2022direct}}: this approach comprises a two-phases training process: an initial coarse training spanning $5$k iterations, followed by a fine training of $10$k iterations, intended to enhance the capability in grasping intricate scene details. We use a batch size of $8192$, maintaining the default scene size.
\end{itemize}

\begin{figure}[t]
    \centering
    \includegraphics[page=4, width=\linewidth]{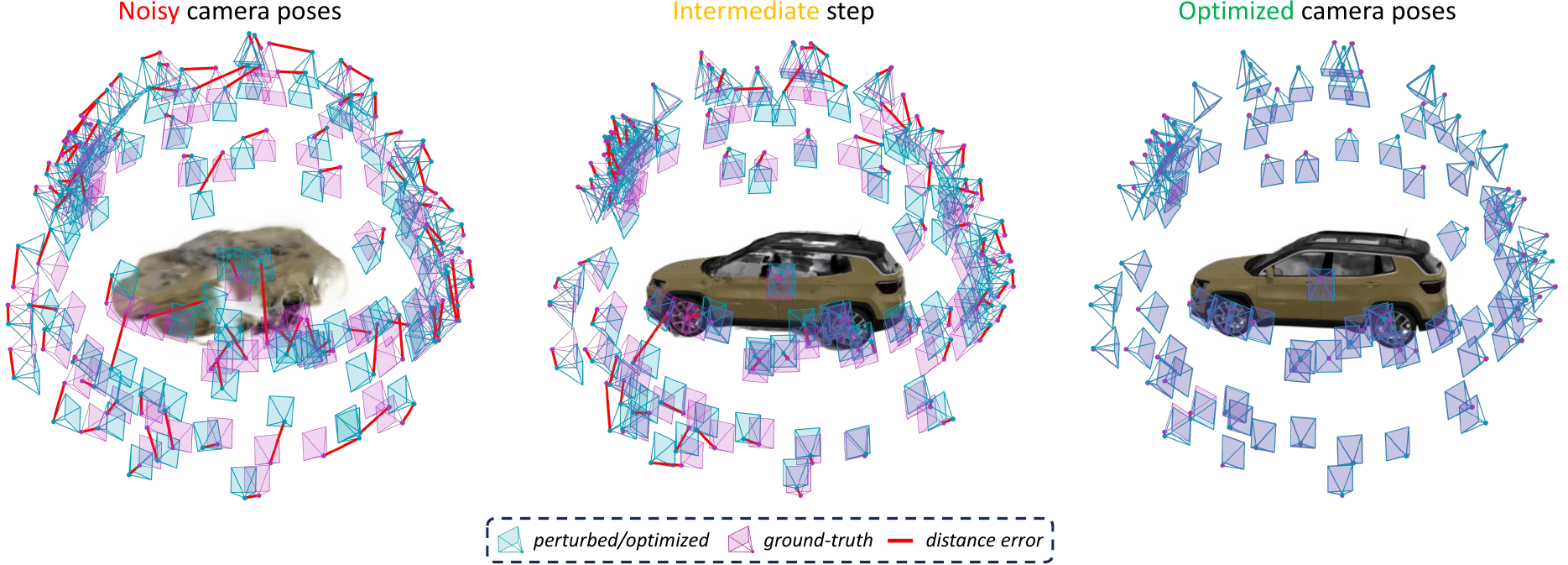}
    \caption{Camera arrangement starting from the noisy initialization (left) to the final \method prediction (right). Note how cameras align with ground-truth at the end.}
    \label{fig:camera_keypoint_optimization}
    \vspace{.25cm}
\end{figure}

\definecolor{myrowcolor}{HTML}{F5F5F5}

\begin{table}[h]
\vspace{-.5cm}
\begin{minipage}[t]{0.49\linewidth}
\centering
    \caption{\method results on CarPatch by varying rotation and translation noise.}
    \resizebox{\linewidth}{!}{

    \begin{tabular}{cc c|c cc c|c ccc}
        \toprule
        $\sigma_{R} (^{\circ})$ & $\sigma_{\mathbf{t}}$ (cm) & & & $\epsilon_{R} (^{\circ})~\downarrow$ & $\epsilon_{\mathbf{t}}$ (cm)~$\downarrow$ & & & PSNR~$\uparrow$ & SSIM~$\uparrow$ & LPIPS~$\downarrow$   \\
         \midrule
         5 & $0.7 \times 10^2$ & & & 0.75 & 3.07& & & 34.34 & 0.982 & 0.014 \\
         \rowcolor{myrowcolor}
         5 & $1.5 \times 10^2$ & & & 1.35 & 3.10 & & & 34.34 & 0.982 & 0.014 \\
         6 & $2.0 \times 10^2$ & & & 3.79 & 3.16 & & & 34.26 & 0.982 & 0.014 \\
         \rowcolor{myrowcolor}
         6 & $2.5 \times 10^2$ & & & 2.34 & 2.13 & & & 34.16 & 0.981 & 0.014 \\
         7 & $3.0 \times 10^2$ & & & 6.54 & 6.21 & & & 33.66 & 0.979 & 0.017 \\

         \bottomrule
    \end{tabular}}
    \label{tab:tab_ablation_synt_noise}
    
\end{minipage}
\begin{minipage}[t]{0.49\linewidth}
\centering
    \caption{\method performance on CarPatch using 2D loss, 3D loss, or both.}

    \resizebox{\linewidth}{!}{
    
        \begin{tabular}{l c|c cc c|c ccc}
        \toprule
         \textbf{Loss} & & & $\epsilon_{R} (^{\circ})~\downarrow$ & $\epsilon_{\mathbf{t}}$ (cm)~$\downarrow$ && & PSNR~$\uparrow$ & SSIM~$\uparrow$ & LPIPS~$\downarrow$ \\
         \midrule
         2D Loss & && 0.75 & 3.10 && & 33.93 & \textbf{0.981} & 0.144 \\
         \rowcolor{myrowcolor}
         3D Loss && & 0.68 & 3.19 && & 33.95 & \textbf{0.981} & \textbf{0.140} \\
         
         (2D+3D) Loss && & \textbf{0.65} & \textbf{3.06} && & \textbf{34.48} & \textbf{0.981} & \textbf{0.140} \\
         \bottomrule
    \end{tabular}}
    \label{tab:tab_ablation_losses}
\end{minipage}
\vspace{-.5cm}

\end{table}

\tit{Results}
As shown in Table~\ref{tab:tab_synt_results}, \method is highly beneficial for 3D reconstruction architectures in synthetic scenarios.  In terms of view synthesis quality metrics (PSNR, SSIM, LPIPS), all the selected baselines outperform Barf and L2G when using \method optimized poses, almost closing the gap with the visual quality obtained by training on ground truth poses. In terms of camera registration quality, relative rotation error increases by $\sim$30\%, while relative translation error decreases by $\sim$42\% compared to L2G. Both \method and L2G demonstrate superior performance compared to Barf. 
The overall alignment achieved by \method closely approximates the ground truth camera poses, as visually depicted in Fig.~\ref{fig:camera_keypoint_optimization}. Moreover, while the additional overhead due to \method over the downstream novel view synthesis can be accurately quantified (30 seconds on a single GPU), the cost for camera registration on Barf/L2G can not be exactly assessed, since radiance fields and cameras are optimized together.

\tit{Additional analysis}
We examine the robustness of our method in synthetic scenarios by introducing varying levels of noise to ground truth camera poses. This was accomplished by altering the normal distribution standard deviation used to randomly sample rotation and translation noise, $\sigma_{R}$ and $\sigma_{\mathbf{t}}$, before adding it to the cameras. As shown in Table \ref{tab:tab_ablation_synt_noise}, results do not show significant deterioration even with a $7^{\circ}$, 3 m noise magnitude. Moreover, as explained in section \ref{subsec:keypoints_3d_to_2d}, \method training loss is made up of two different components: one operating on the 2D image plane and the other in the 3D common space. Table~\ref{tab:tab_ablation_losses} shows that their combination further improves performance compared to using them individually. All experiments adopt Gaussian Splatting~\cite{kerbl20233d}.

\definecolor{myrowcolor}{HTML}{F5F5F5}

\begin{table*}[t]
    \centering
        \caption{Quantitative results on the KRONC dataset. The GaussianSplatting~\cite{kerbl20233d} baseline trained with COLMAP poses and an optimized standard trajectory. The results in (·) are computed after masking out the background.}
    \resizebox{\linewidth}{!}{
    \begin{tabular}{ll cc ccc ccc}
        \toprule
        & & & & \multicolumn{3}{c}{\textbf{Full scene(Masked vehicle)}}  \\
        [0.35em]
         \textbf{Env} & \textbf{Vehicle} & \textbf{Init Pose} & \textbf{\# Opt. Cameras} & PSNR~$\uparrow$ & SSIM~$\uparrow$ & LPIPS~$\downarrow$  \\
         \midrule
         Env1 & Ford-Focus & COLMAP & 161/161 & 29.11 (28.37) & 0.916 (0.959) & 0.089 (0.036) \\
         \rowcolor{myrowcolor}
         Env1 & Fiat-500L & COLMAP & 143/143 & 26.94 (25.12) & 0.892 (0.930) & 0.115 (0.061)  \\
         Env1 & Hyundai-i10 & COLMAP & 123/123 & 29.38 (29.45) & 0.918 (0.971) & 0.097 (0.026)\\
         \rowcolor{myrowcolor}
         Env2 & Fiat-500L & COLMAP & 94/94 & 28.15 (28.60) & 0.922 (0.945) & 0.117 (0.048)  \\
         Env2 & Toyota-Yaris & COLMAP & 91/91 & 28.90 (29.18) & 0.936 (0.957) & 0.115 (0.029)  \\
         \rowcolor{myrowcolor}
         Env3 & Toyota-Yaris & COLMAP & 116/116 & 31.00 (33.31) & 0.948 (0.983) & 0.065 (0.017)  \\
         Env3 & Hyundai-i10 & COLMAP & 123/123 & 30.95 (31.11) & 0.942 (0.974) & 0.072 (0.025)  \\
         \midrule
          Env1 & Ford-Focus & Trajectory & 161/161
          & 21,97 (23,41) & 0.696 (0,888) & 0.296 (0.081) \\
         \rowcolor{myrowcolor}
         Env1 & Fiat-500L & Trajectory & 124/143 
         & 20.52 (21.56) & 0.652 (0.835) & 0.318 (0.121)  \\
         Env1 & Hyundai-i10 & Trajectory & 121/123 
         & 21.46 (23.38) & 0.666 (0.896) & 0.296 (0.070)  \\
         \rowcolor{myrowcolor}
         Env2 & Fiat-500L & Trajectory & 67/94  
         & 16.94 (19.20) & 0.660 (0.769) & 0.359 (0.176)  \\
         Env2 & Toyota-Yaris & Trajectory & 90/91 
         & 17.68 (21.54) & 0.727 (0.836) & 0.348 (0.125)  \\
         \rowcolor{myrowcolor}
        Env3 & Toyota-Yaris & Trajectory & 116/116
         & 19.06 (21.23) & 0.601 (0.850) & 0.396 (0.130)  \\
        Env3 & Hyundai-i10 & Trajectory & 107/123
         & 18.27 (22.88) & 0.582 (0.892) & 0.405 (0.091)  \\
         \bottomrule
    \end{tabular}
    }
    \vspace{-.5cm}
    \label{tab:tab_results_real}
\end{table*}

\subsection{Real-world vehicle scenes}

\label{subsec:experiments_real}
To assess the performance of our method in the real domain, we use the proposed \dataset as our benchmark. As described in Sec.~\ref{sec:dataset}, semantic keypoints information come from~\cite{kreiss2021openpifpaf}.

\tit{Implementation and experimental settings} Differently from the synthetic scenario, no ground truth camera poses are available in the \dataset. In this case, we run the COLMAP algorithm on each scene to retrieve a pseudo-ground truth to be used as our reference. Driven by what usually happens in real contexts and considering a reasonable dimension of the scene, we define a standard $4$m radius circular trajectory, placing as many cameras as the number of vehicle images, forward-facing and with no tilt angle. We refer to this trajectory as our initial coarse camera configuration (the same for all real scenes), which we optimize using \method. We follow the LLFF dataset~\cite{mildenhall2019llff} train/test split protocol sampling one test image every $8$ frames for each recording. 
We select Gaussian Splatting~\cite{kerbl20233d} as the 3D reconstruction baseline based on the results obtained on the synthetic scenario. Experiments are conducted with the same configuration described in Sec.~\ref{subsec:experiments_synthetic}, with an image resolution of $480\times270$.

\tit{Results} In Table~\ref{tab:tab_results_real}, we assess the performance of our algorithm with respect to COLMAP camera registration. As a reference, the maximum PSNR achieved by training Gaussian Splatting with the initial coarse trajectory is 12.0 on the \textit{Ford-Focus} scene. Bundle-adjustment methods (like L2G) are not able to converge in this inward-facing 360° setting with large rotations, as mentioned in their paper~\cite{chen2023local} and demonstrated by our preliminary experiments (starting from both identity transformation and our circular trajectory). L2G obtains a PSNR lower than 10.0 for all the \dataset scenes.
\method is able to find a reasonable camera configuration, reaching a maximum PSNR of 23.41 on the \textit{Ford-Focus} scene (with masked out background).
We test the visual quality of the reconstruction using both full images and masked backgrounds with the Gaussian Splatting baseline. The performance drop compared to COLMAP is partly due to the keypoint detector recall, which may leave some viewpoints without keypoint annotations, causing those poses to remain unadjusted by \method.\definecolor{myrowcolor}{HTML}{F5F5F5}
\begin{wraptable}{r}{0.45\linewidth}
    \vspace{-1.1cm}

    \centering
        \caption{\method results with varying depth initialization on \dataset for masked cars (* indicates unoptimized depth).}
        \vspace{.25cm}
        \resizebox{\linewidth}{!}{
    \begin{tabular}{l ccc}
        \toprule
         \textbf{Depth type} & PSNR~$\uparrow$ & SSIM~$\uparrow$ & LPIPS~$\downarrow$ \\
         \midrule
         DinoV2~\cite{oquab2023dinov2}$^*$ & 17.54 & 0.703 & 0.223 \\
         \rowcolor{myrowcolor}
         DinoV2~\cite{oquab2023dinov2} & 20.53 & 0.827 & 0.149 \\
         Random  & \textbf{21.89} & \textbf{0.852} & \textbf{0.114} \\
         \bottomrule
    \end{tabular}}
    \label{tab:tab_ablation_depth_real}
    \vspace{-.7cm}
\end{wraptable} In particular, this can be noted in the Env2 \textit{Fiat-500L} scene, which has only 13 keypoints per image on average (according to Tab.~\ref{tab:tab_kronc_dataset}), leading to almost $30\%$ of the camera viewpoints being discarded in the optimization process. Even if the performance gap is noticeable, \method is $\sim$16 times faster than COLMAP, \ie~30 seconds \vs 8 minutes on a single GPU for the same number of images. It is worth noting that this comparison does not take into account the inference time needed for OpenPifPaf~\cite{kreiss2021openpifpaf} keypoint extraction, which is 33 seconds on average over the \dataset scenes. This leads to an effective $\sim8\times$ speedup.

\begin{figure}[t]
    \centering
    \includegraphics[width=0.8\linewidth]{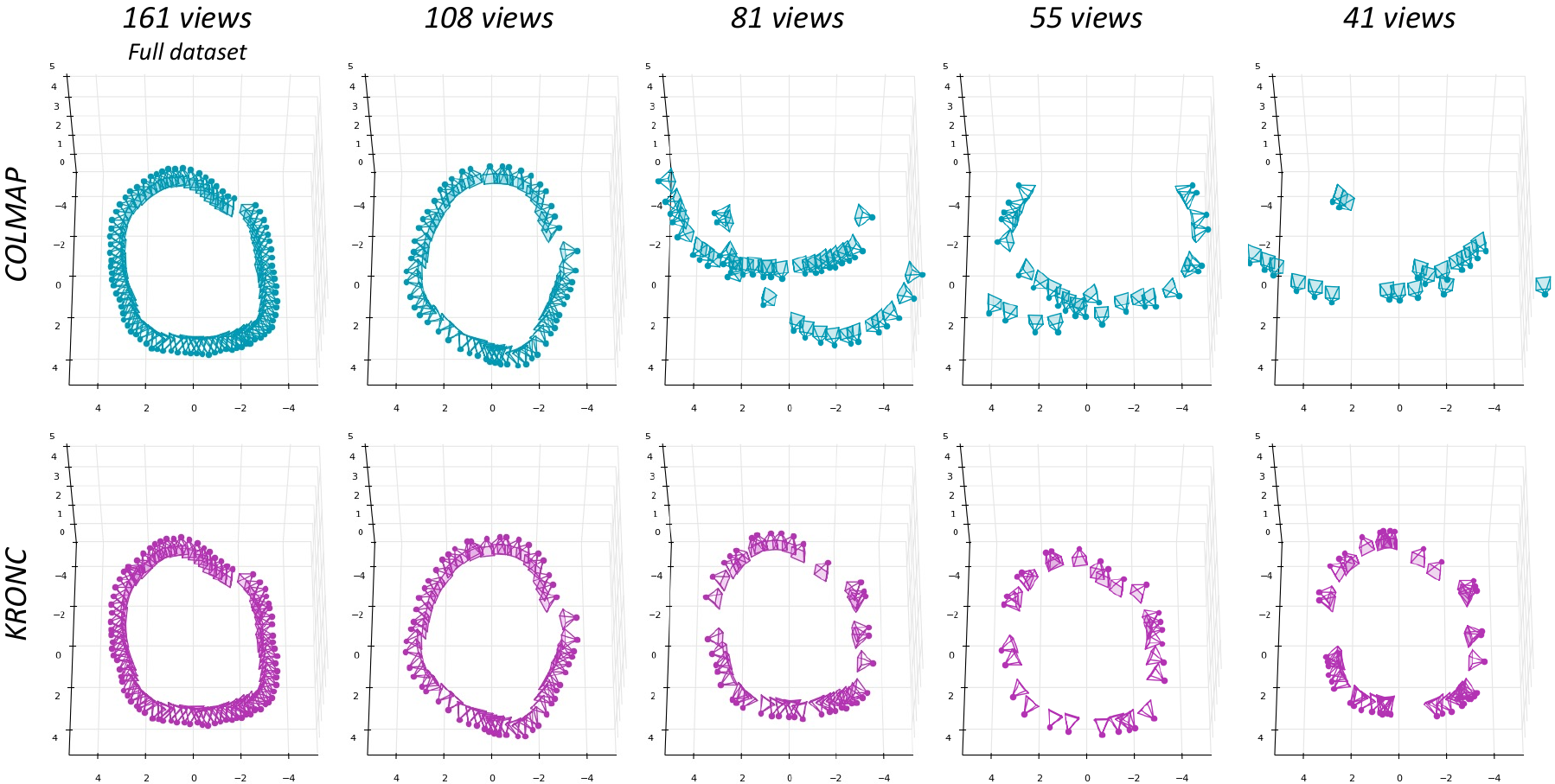}
    \caption{Comparison between COLMAP and \method for camera pose reconstruction on the \dataset's \textit{Ford-Focus} using different subsets of the original full scene.}
    \label{fig:trajectories_real}
    \vspace{-.5cm}
\end{figure}

\tit{Additional analysis} For real scenes, the $z_i^j$ depth values are randomly initialized following the same approach described in Sec.~\ref{subsec:experiments_synthetic} for synthetic scenes. Here we investigate the impact of depth initialization by considering all the \dataset.
As an alternative, we provide results by initializing depths using predictions from DinoV2~\cite{oquab2023dinov2}. In a preliminary experiment, these depths are not further optimized within the \method iterations, and are kept fixed. In a second scenario, we subsequently refine depths during the \method optimization process. As shown in Table~\ref{tab:tab_ablation_depth_real}, the random initialization based on the scene scale obtains the best results in all the visual metrics.%, even though additional fine-tuning improves results over freezed DinoV2 depths.

Finally, we assess the robustness of the \method algorithm in a real-world scenario by sub-sampling the number of images used from the \textit{Ford-Focus} scene within the \dataset. As illustrated in Fig.~\ref{fig:trajectories_real}, COLMAP camera pose estimation capability rapidly degrades when reducing the number of images (\ie~decreasing image overlap), as already noted in~\cite{truong2023sparf}. In contrast, \method results demonstrate that by replacing pairwise matches with global reasoning via shared semantic keypoints and by coarsely initializing camera poses using some prior knowledge, robust registration can be achieved even with limited data.

\subsection{Qualitative results}
\begin{figure}[t]
    \centering
    \includegraphics[page=2,width=0.93\linewidth]{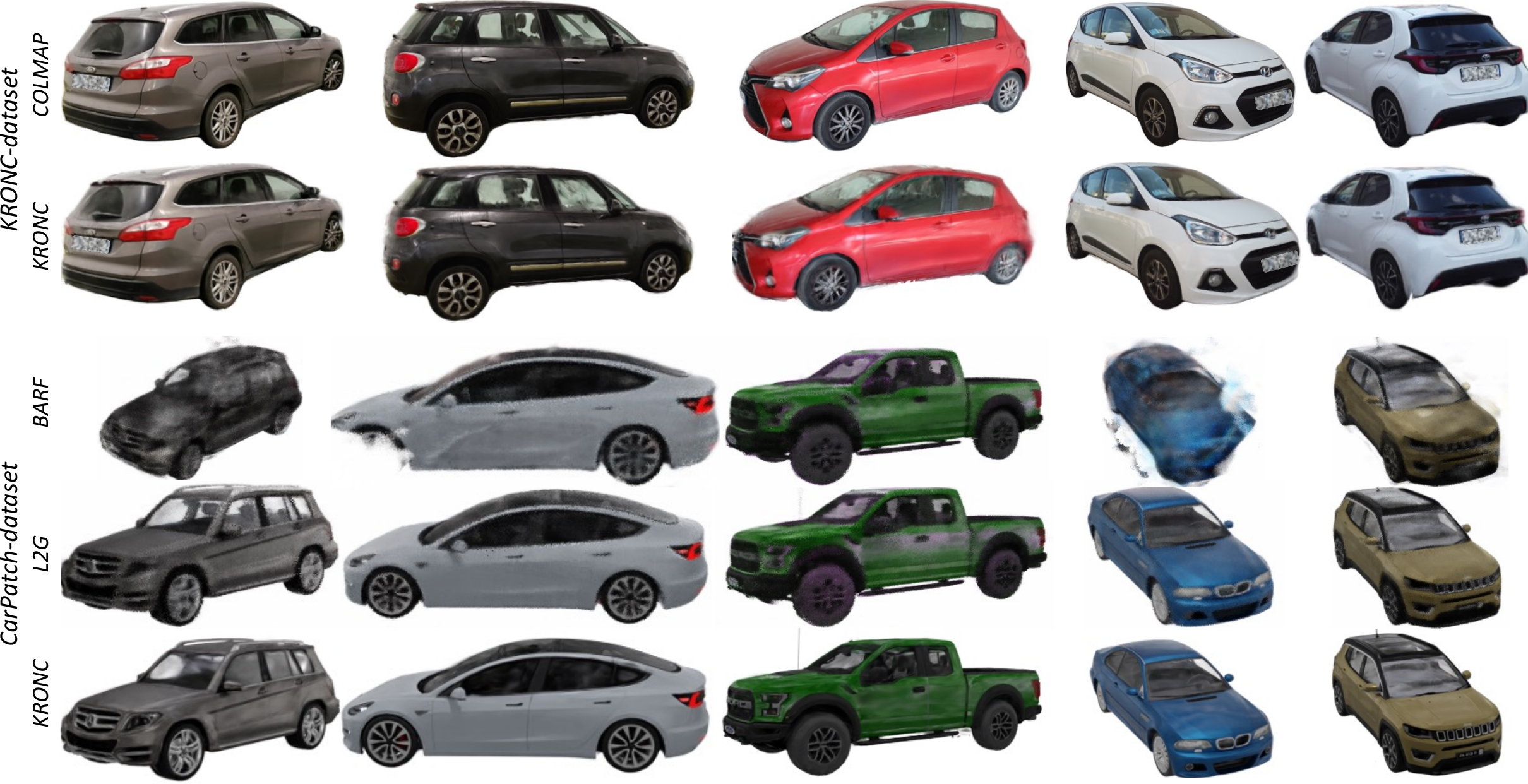}
    \caption{Qualitative results of \method followed by Gaussian Splatting on real scenes (first two rows) and synthetic ones (last three rows). Best viewed in color and zoom. }
    \label{fig:qualitative_real}
    \vspace{-.5cm}
    \end{figure}

In Fig.~\ref{fig:qualitative_real} we show some qualitative samples obtained with the Gaussian Splatting baseline after \method camera optimization in both the synthetic and real scenarios. The proposed method is able to recover a camera configuration to obtain a high-quality reconstruction of the synthetic vehicles. Also in the more challenging real scenario \method confirms its robustness finding a consistent sub-optimal camera configuration for a realistic 3D vehicle reconstruction.

\section{Conclusion}
\label{sec:conclusion}
We presented both a new dataset and a state-of-the-art algorithm to foster research and applications on the \textit{vehicle inspection} task. The \dataset represents the first collection of high-quality scenes of real vehicles, while the \method algorithm specifically tackles camera optimization using 2D keypoints as a pre-processing step for novel view synthesis. With almost no overhead, \method efficiently recovers camera poses, yielding reconstruction results comparable to those obtained with ground truth cameras for synthetic scenes. Similar observations have been demonstrated on the real scenes from the \dataset, by only assuming an initial circular trajectory of the cameras. Despite the advantages of the \method algorithm \wrt S\textit{f}M and bundle-adjusting novel view synthesis approaches, it still has some limitations. Its performance on real-world scenes highly depends on the quality of predicted keypoints, when extracted with an automatic detection method, as demonstrated in Sec.~\ref{subsec:experiments_real}. Moreover, it needs at least a rough initialization of the camera poses, being not able to converge to a good solution when starting from random values.

\bibliographystyle{splncs04}
\bibliography{biblio}

\newpage
\appendix

\section{Reproducibility}
%For the sake of reproducibility, we release a demo code together with two annotated scenes, one from the augmented CarPatch~\cite{di2023carpatch} dataset and one from the new \dataset (downsampled by 0.25 due to zip dimension restriction on CMT).  This demo optimizes camera poses and visually shows how reprojected keypoint cluster centroids converge to the corresponding keypoint locations on the image planes as a consequence of the \method optimization. Upon publication, we will release the complete \dataset together with detailed instructions for training all the considered novel view synthesis baselines with camera extrinsics estimated by \method.

Upon publication, we will release the complete \dataset together with detailed instructions for training all the considered novel view synthesis baselines with camera extrinsics estimated by \method.

\section{Additional details}
\label{sec:impl-details}

\tit{Implementation details}
%Since we consider all the images in a scene being acquired with the same sensor and resolution, the intrinsic camera parameters are initialized using the maximum value between the image's width and height. The principal point of the camera is set to the center of the image plane. 
The extrinsic parameters are optimized by disentangling rotation and translation. Since rotation and translation noises have different effects on vehicle visibility, optimizing both parameters in the same way is not trivial. 
All experiments on both the main paper and this supplementary material have been run on a machine with an Intel Core i7-12700F and a NVIDIA GeForce GTX 1080 Ti. With this hardware configuration, the \method algorithm runs $10$K iterations in $30$ seconds on GPU. We use the Adam optimizer with a learning rate of 0.01 for the synthetic data and 0.001 for the real data. We apply a cosine annealing decay with a decay factor of $0.001$. Being $N$ the number of views for a scene and $J$ the number of semantic keypoints, we optimize a 6D vector and a 3D vector for rotation and translation for each view. Moreover, a vector of $J$ keypoint depths is optimized, leading to a total of 9$N$+$JN$ parameters. Considering a scene with 100 views and 66 keypoints, the \method algorithm optimizes only 7.5K parameters, making it suitable even for edge devices.

\tit{Camera noise}
In all the synthetic scenario experiments, we introduce perturbations to the ground truth camera poses using additive noise. It's noteworthy that our strategy for adding noise differs from Barf~\cite{lin2021barf}, where ground-truth camera poses are perturbed using left multiplication, transforming cameras around the object's center. In this setting, the transformed cameras maintain their orientation toward the object's center, and the distances between the cameras and the object are not largely modified.

In contrast, our approach follows the perturbation strategy proposed by L2G-Nerf~\cite{chen2023local}, which involves perturbing ground-truth camera poses using right multiplication, transforming cameras around themselves. This perturbation affects both camera viewing directions (which may not always face the object's center) and camera positions, consequently altering the distances between the cameras and the object.

\begin{figure}[t]
     \centering
     \includegraphics[page=2,width=\linewidth]{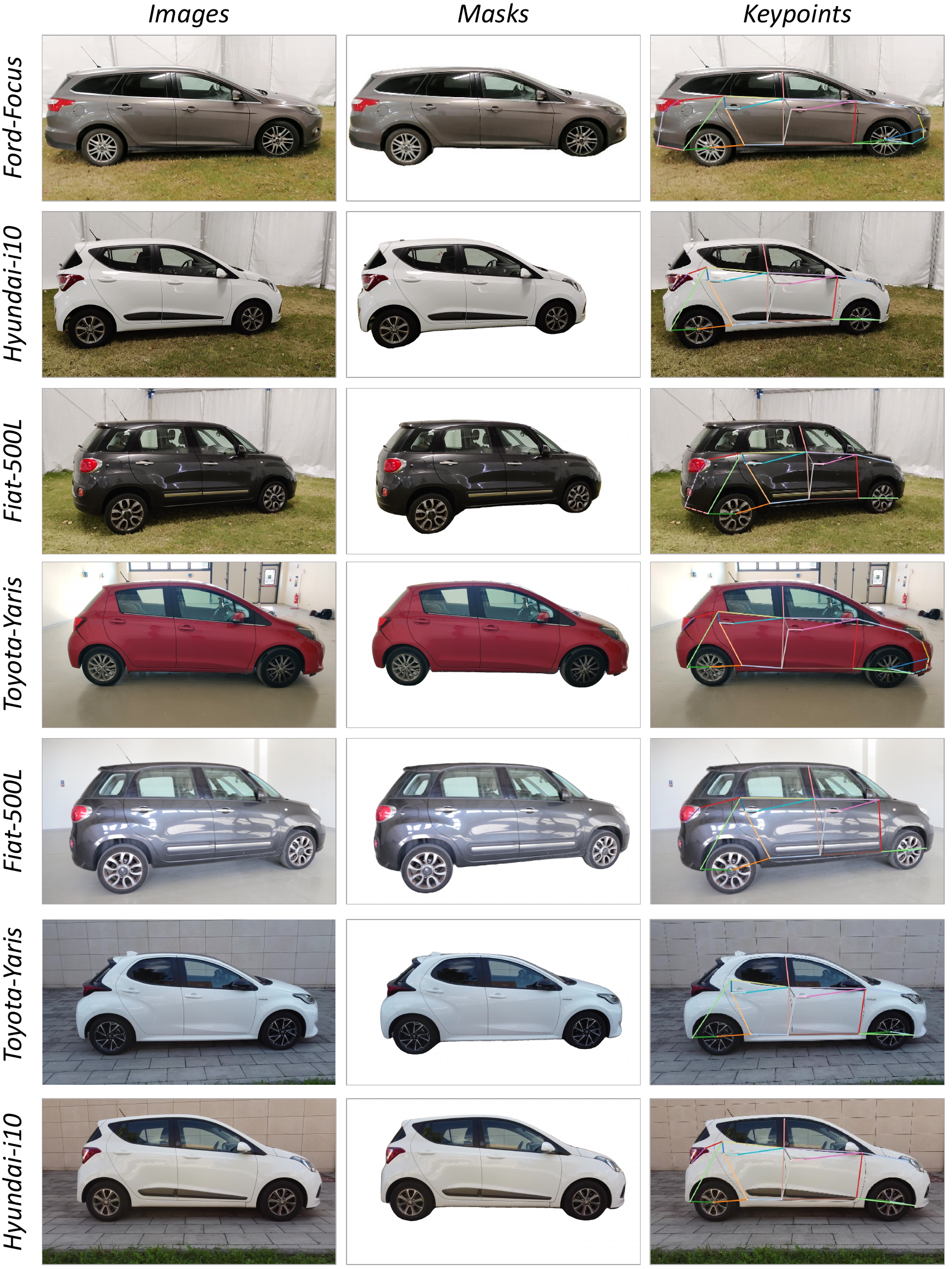}
     \caption{Overview of the \dataset showing the full-scene images, the segmented vehicles, and the predicted keypoints.}
     \label{fig:figura_dataset}
     \vspace{-.5cm}
\end{figure}

\tit{Dataset}
As described in Section \textcolor{red}{3} of the main paper, our dataset captures a diverse set of 7 vehicles across 3 distinct environments. Figure \ref{fig:figura_dataset} showcases example captures from each environment, along with keypoint and mask annotations.

\section{Additional quantitative results}
The CarPatch~\cite{di2023carpatch} dataset provides ground-truth camera pose annotations, which can be thought of as an upper bound for \method optimization, as already done in Sec. \textcolor{red}{5.1} of the main paper. In this section, we show additional ablation studies performed on the synthetic data.

\input{tables/supp_result_synt_all_methods}

\definecolor{myrowcolor}{HTML}{F5F5F5}

\begin{table}[t]
    \centering
    \caption{\method performances by sampling a different number of poses from the CarPatch dataset.}
    \resizebox{0.55\linewidth}{!}{
    \begin{tabular}{c c|c cc c|c ccc}
        \toprule
         \textbf{\# poses} & & & \cellcolor{white} $\epsilon_{R} (^{\circ})$~$\downarrow$  & \cellcolor{white} $\epsilon_{\mathbf{t}}$ (cm)~$\downarrow$ && & PSNR~$\uparrow$ & SSIM~$\uparrow$ & LPIPS~$\downarrow$ \\
         \midrule
         5 & && 1.72 & 3.37 && & 20.62 & 0.890 & 0.092 \\
         \rowcolor{myrowcolor}
         10 && & 0.73 & 3.31 && & 24.86 & 0.929 & 0.052 \\

         20 && & 1.62 & 3.24 && & 29.88 & 0.958 & 0.028 \\
         \rowcolor{myrowcolor}
         30 && & 2.24 & 3.12 && & 31.50 & 0.967 & 0.023 \\
         40 && & 1.82 & 3.10 && & 32.59 & 0.974 & 0.019 \\
         \rowcolor{myrowcolor}
         50 && & 1.18 & 3.07 && & 33.33 & 0.977 & 0.017 \\
         60 && & 0.93 & 3.06 && & 33.34 & 0.977 & 0.017 \\
         \rowcolor{myrowcolor}
         70 && & 0.69 & 3.07 && & 34.15 & 0.981 & 0.014 \\
         \bottomrule
    \end{tabular}
    }
    \label{tab:tab_ablation_n_poses}
\end{table}
\definecolor{myrowcolor}{HTML}{F5F5F5}
\definecolor{first}{HTML}{dec546}
\definecolor{second}{HTML}{d7d7d7}
\definecolor{third}{HTML}{7e4205}

\begin{table*}[t]
\centering
\caption{Performance comparison of \method + GaussianSplatting and L2G-NeRF on CarPatch dataset with different noise levels.}
\resizebox{\linewidth}{!}{

\begin{tabular}{ c c | c c c c c | c c c c c}
    \toprule 
        & &  \multicolumn{5}{c|}{KRONC+GaussianSplatting} & \multicolumn{5}{c}{L2G-NeRF} \\
        \cellcolor{white} $\sigma_{R} (^{\circ})$ & \cellcolor{white} $\sigma_{\mathbf{t}}$ (cm) &  \cellcolor{white} $\epsilon_{R} (^{\circ})~\downarrow$ & \cellcolor{white} $\epsilon_{\mathbf{t}}$ (cm)~$\downarrow$ & \cellcolor{white} PSNR~$\uparrow$ & SSIM~$\uparrow$ & \cellcolor{white} LPIPS~$\downarrow$ &  \cellcolor{white} $\epsilon_{R} (^{\circ})~\downarrow$ & \cellcolor{white} $\epsilon_{\mathbf{t}}$ (cm)~$\downarrow$ & \cellcolor{white} PSNR~$\uparrow$ & SSIM~$\uparrow$ & \cellcolor{white} LPIPS~$\downarrow$ 
       \\
       
    \midrule
    
     \cellcolor{white} 5 & \cellcolor{white} $0.7 \times 10^2$
    & \cellcolor{white} 0.75 
    & \cellcolor{white} 3.07 
    & \cellcolor{white} 34.34 
    & \cellcolor{white} 0.982 
    & \cellcolor{white} 0.014 
    & \cellcolor{white} 0.51
    & \cellcolor{white} 6.25
    & \cellcolor{white} 31.64
    & \cellcolor{white} 0.965
    & \cellcolor{white} 0.062
    \\
    
    \cellcolor{myrowcolor} 5 & \cellcolor{myrowcolor} $1.5 \times 10^2$& \cellcolor{myrowcolor} 1.35 & \cellcolor{myrowcolor} 3.10 & \cellcolor{myrowcolor} 34.34 & \cellcolor{myrowcolor} 0.982 & \cellcolor{myrowcolor} 0.014 
    & \cellcolor{myrowcolor} 8.19
    & \cellcolor{myrowcolor} 78.0
    & \cellcolor{myrowcolor} 18.51
    & \cellcolor{myrowcolor} 0.876
    & \cellcolor{myrowcolor} 0.129
    \\
    
    \cellcolor{white} 6 & \cellcolor{white} 
    $2.0 \times 10^2$ & \cellcolor{white} 3.79  & \cellcolor{white} 3.16  & \cellcolor{white} 34.26  & \cellcolor{white} 0.982
    & \cellcolor{white} 0.014 
    & \cellcolor{white} 14.38 
    & \cellcolor{white} 177
    & \cellcolor{white} 15.64
    & \cellcolor{white} 0.844
    & \cellcolor{white} 0.171  \\
    
    \cellcolor{myrowcolor} 6 & \cellcolor{myrowcolor} $2.5 \times 10^2$  & \cellcolor{myrowcolor} 2.34  & \cellcolor{myrowcolor} 2.13  & \cellcolor{myrowcolor} 34.16  & \cellcolor{myrowcolor} 0.981 & \cellcolor{myrowcolor} 0.014  & \cellcolor{myrowcolor} 24.39
    & \cellcolor{myrowcolor} 269
    & \cellcolor{myrowcolor} 12.02
    & \cellcolor{myrowcolor} 0.793
    & \cellcolor{myrowcolor} 0.252 \\
        
  \cellcolor{white} 7 & \cellcolor{white} $3.0 \times 10^2$  & \cellcolor{white} 6.54  & \cellcolor{white} 6.21  & \cellcolor{white} 33.66  & \cellcolor{white} 0.979  & \cellcolor{white} 0.017  
    & \cellcolor{white} 31.29
    & \cellcolor{white} 348
    & \cellcolor{white} 11.19
    & \cellcolor{white} 0.778
    & \cellcolor{white} 0.267
    \\
    
    \bottomrule
    
\end{tabular}
\label{tab:tab_noise_level_all_methods}
}
\vspace{-0.3cm}
\end{table*}

\tit{KRONC vs state-of-the-art}
Table \ref{tab:tab_synt_all_results} comprehensively details the performance of our method compared to the state-of-the-art on each scene of the CarPatch dataset.
Our proposed method achieves performance comparable to L2G-NeRF in terms of rotation and translation metrics, while simultaneously establishing state-of-the-art results on PSNR, SSIM, and LPIPS metrics when combined with Gaussian Splatting.

\tit{Number of training poses}
In Table \ref{tab:tab_ablation_n_poses}, we show \method's robustness by varying the number of training views, keeping test views unaltered. Given a number of training views, results are averaged over all the scenes with that specific number of views. Our results showcase the method's capability to refine noisy poses even with limited data, leading to performance gains as the number of cameras increases.

\tit{Different noise levels}
Our method, combined with Gaussian Splatting, demonstrates superior robustness to noise compared to the L2G-NeRF architecture, as shown in Table~\ref{tab:tab_noise_level_all_methods}.  While our method maintains accurate rotation and translation estimates across all noise levels tested, L2G-NeRF fails to reconstruct camera positions accurately when the translation noise exceeds 70cm. 

\begin{figure}[t]
     \centering
     \includegraphics[page=1,width=\linewidth]{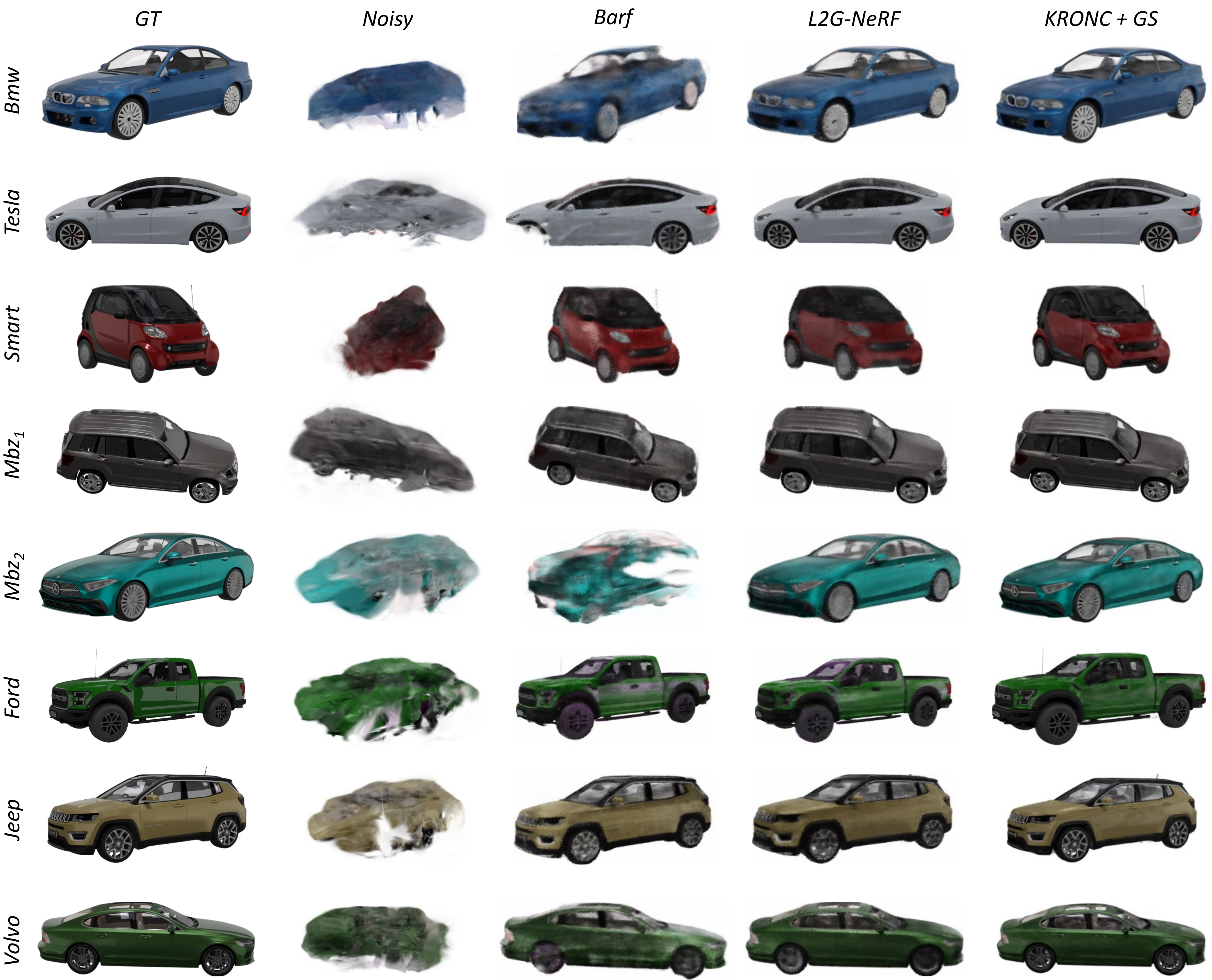}
     \caption{Comparison of qualitative results across all scenes in the CarPatch dataset, showcasing vehicle reconstructions from Barf, L2G-NeRF, and KRONC + Gaussian Splatting.}
     \label{fig:qualitative_synt}
\end{figure}

\section{Additional qualitative results}
In this section, we show a qualitative comparison with respect to state-of-the-art approaches in the synthetic scenario. In the real-world scenes, we compare the quality of the reconstruction obtained with coarse or optimized camera trajectories.

\tit{\method vs state-of-the-art}
Figure~\ref{fig:qualitative_synt} presents a qualitative comparison among various methods utilized for reconstructing vehicles in the CarPatch dataset from noisy camera poses. Barf encounters challenges in accurately reconstructing vehicles, while L2G-NeRF demonstrates greater consistency in this task. Notably, leveraging \method alongside Gaussian Splatting (GS) leads to a more precise vehicle reconstruction, effectively capturing intricate details.

\begin{figure}
     \centering
     \includegraphics[page=1,width=\linewidth]{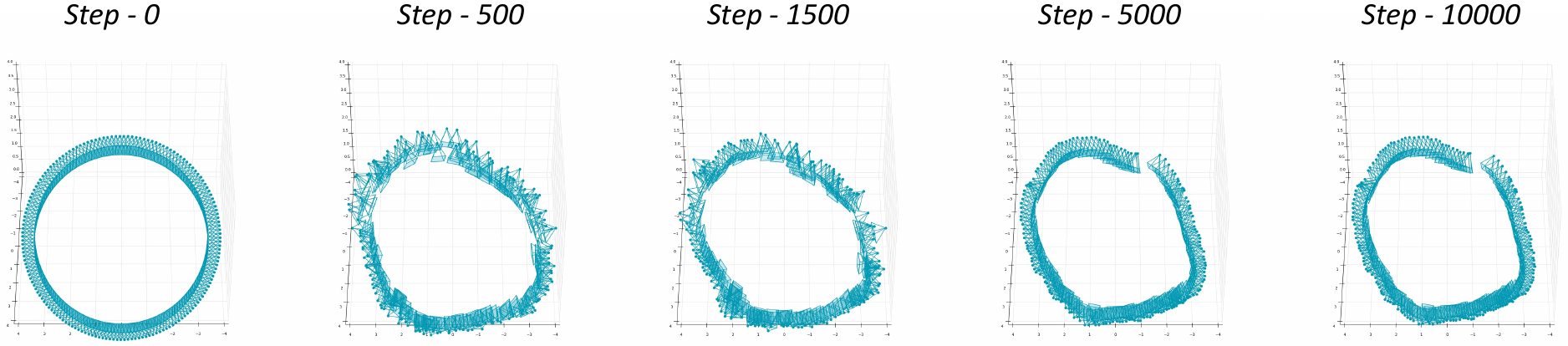}
     \caption{Optimization steps in a real scenario. On the left, a visualization of the initial circular trajectory. On the right, the optimized trajectory at each intermediate step.}
     \label{fig:step_traj}
\end{figure}

\tit{Trajectory optimization}
Figure~\ref{fig:step_traj} illustrates the trajectory optimization process for real-world scenarios, as detailed in Section~\textcolor{red}{5.2} of the main paper. The initial trajectory (left) starts as a generic circular path, which is progressively refined in the following iterations to achieve a reliable and reasonable camera registration (right).
 
\begin{figure}[t]
     \centering
     \includegraphics[page=2,width=\linewidth]{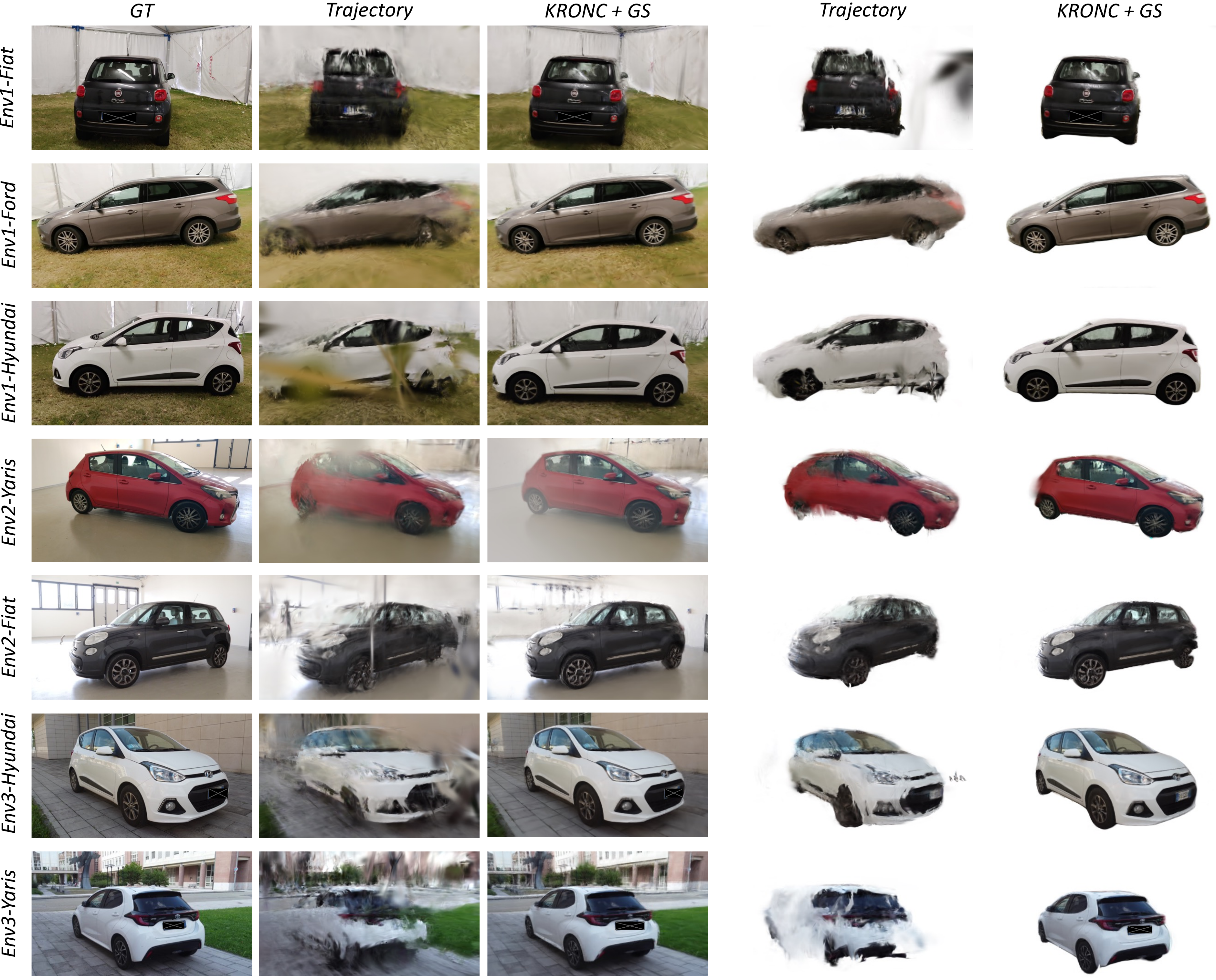}
     \caption{Qualitative results of KRONC + Gaussian Splatting on the KRONC-dataset. The second and third columns showcase reconstructions using coarse and optimized trajectories, while the last two columns display reconstructions utilizing masked images.}
     \label{fig:qualitativi_real_supp}
\end{figure}

\tit{Coarse vs optimized trajectory} In Figure~\ref{fig:qualitativi_real_supp}, we present qualitative results illustrating \method's capability to reconstruct vehicles in a real-case scenario. Starting from the initialization of cameras, as detailed in~\textcolor{red}{5.2} of the main paper, our method successfully achieves an enhanced vehicle reconstruction. This improvement is evident in both environments, with or without background.

\end{document}